\documentclass[runningheads]{llncs}
\usepackage[T1]{fontenc}
\usepackage{graphicx}
\usepackage{booktabs}
\usepackage{multirow}
\usepackage{array} 
\usepackage{tabularx}
\usepackage{subcaption}
\usepackage{graphicx}
\usepackage{float}
\usepackage[most]{tcolorbox}
\begin{document}
\title{From Thinking to Output: Chain-of-Thought and Text Generation Characteristics in\\Reasoning Language Models }
\titlerunning{From Thought to Output: A Comparative Study}

\author{
    Junhao Liu\inst{1}\orcidID{0009-0007-9353-6929} \and
    Zhenhao Xu\inst{1}\orcidID{0009-0002-1834-5181} \and
    Yuxin Fang\inst{1}\orcidID{0009-0006-5220-3274} \and
    Yichuan Chen\inst{1}\orcidID{0009-0006-4992-5559} \and
    Zuobin Ying\inst{1}\orcidID{0000-0002-1658-4931} \and
    Wenhan Chang\thanks{Corresponding Author}\inst{2}\orcidID{0000-0003-3350-5171}
}
\authorrunning{J. Liu et al.}
%
\institute{Faculty of Data Science, City University of Macau, Macau SAR, China \\
\email{611yuyi@gmail.com, \{D23090100776,D24090103371,D24090150413,zbying\}@cityu.edu.mo} \and
School of Information Engineering, Zhongnan University of Economics and Law, Wuhan, China\\
\email{changwh530@gmail.com}}
\maketitle 
\begin{abstract}

Recently, there have been notable advancements in large language models (LLMs), demonstrating their growing abilities in complex reasoning. However, existing research largely overlooks a thorough and systematic comparison of these models' reasoning processes and outputs, particularly regarding their self-reflection pattern (also termed "Aha moment"~\cite{deepseekai2025deepseekr1incentivizingreasoningcapability}) and the interconnections across diverse domains. This paper proposes a novel framework for analyzing the reasoning characteristics of four cutting-edge large reasoning models (GPT-o1~\cite{openai2024openaio1card}, DeepSeek-R1~\cite{deepseekai2025deepseekr1incentivizingreasoningcapability}, Kimi-k1.5~\cite{kimiteam2025kimik15scalingreinforcement}, and Grok-3) using keywords statistic and LLM-as-a-judge paradigm. Our approach connects their internal thinking processes with their final outputs. A diverse dataset consists of real-world scenario-based questions covering logical deduction, causal inference, and multi-step problem-solving. Additionally, a set of metrics is put forward to assess both the coherence of reasoning and the accuracy of the outputs. The research results uncover various patterns of how these models balance exploration and exploitation, deal with problems, and reach conclusions during the reasoning process. Through quantitative and qualitative comparisons, disparities among these models are identified in aspects such as the depth of reasoning, the reliance on intermediate steps, and the degree of similarity between their thinking processes and output patterns and those of GPT-o1. This work offers valuable insights into the trade-off between computational efficiency and reasoning robustness and provides practical recommendations for enhancing model design and evaluation in practical applications. 
We publicly release our project at: \url{https://github.com/ChangWenhan/FromThinking2Output}

\keywords{Reasoning Language Models, Reasoning Process, Reasoning Pattern Analysis}
\end{abstract}

\section{Introduction}


In recent years, large language models (LLMs)~\cite{zhao2025surveylargelanguagemodels} have witnessed remarkable development and have demonstrated their significant potential in various natural language processing tasks. Their ability to perform complex reasoning has become crucial, enabling applications such as intelligent question-answering, decision-making support, and logical analysis. For instance, in information retrieval~\cite{DBLP:conf/coling/ZhangZD25}, LLMs can help users find relevant information more accurately by understanding complex queries and inferring the underlying intentions. In the medical domain, they can assist doctors in diagnosing diseases by analyzing patient symptoms and medical records through logical reasoning~\cite{liu2025generalist,zhou2024largelanguagemodelsdisease}.

With the continuous improvement of LLMs, researchers have been exploring their reasoning capabilities. Some studies have focused on altering the behaviors of models through techniques such as fine-tuning on specific datasets~\cite{chang2024classmachineunlearningcomplex}. Others have investigated the interpretability~\cite{chang2024zeroshotclassunlearninglayerwise} of the reasoning process in LLMs~\cite{huang-chang-2023-towards,mondorf2024accuracyevaluatingreasoningbehavior}, aiming to understand how models arrive at their conclusions. However, most of these studies either improve individual models or analyze a single model's reasoning process.

Despite progress, there are still notable drawbacks in the current research on LLMs' reasoning. A major limitation is the lack of systematic comparisons among different models' reasoning processes and outputs. Most existing studies do not comprehensively evaluate how different models handle various reasoning tasks and how their internal thinking mechanisms vary. Without such comparisons, it is difficult to fully understand the strengths and weaknesses of different LLMs in reasoning, and it is also challenging to optimize model design and training strategies.

Our research takes a comprehensive approach to addressing these issues. First, we construct a dataset encompassing eight real-world domains: finance, law, and mathematics. This dataset contains many questions requiring logical deduction, causal inference, and multi-step problem-solving. Using this diverse dataset, we aim to simulate various real-world reasoning scenarios.

Then, we conduct a detailed comparison of the reasoning processes of different reasoning language models. Specifically, we analyze the differences in the number of reflections and the keywords used during the reflection process among these models. This helps us understand how different models approach problem-solving and adjust their reasoning strategies. In addition, we examine the similarity between the reasoning processes of different models and that of GPT-o1, a well-known and powerful LLM. We can identify the unique characteristics and commonalities of different models' thinking processes by comparing the step-by-step reasoning paths.

Furthermore, we also evaluate the similarity between the output contents of different models and that of GPT-o1. This includes not only the answers' accuracy but also the responses' structure and style. Through these comparisons, we can gain insights into how different models generate outputs and how they differ from a benchmark model.

Our research makes several significant contributions: 
\begin{enumerate}
    \item By proposing a novel framework that links the internal thinking processes of LLMs to their final outputs, we provide a new perspective for studying and comparing different models. This study can be a valuable tool for future research in this area.
    \item Our detailed analysis of the differences in various models' reasoning processes and outputs reveals distinct patterns of how models balance exploration and exploitation, handle problems and draw conclusions. These findings contribute to a deeper understanding of the reasoning mechanisms of LLMs.
    \item By comparing the models in the way we do, we can infer the distribution differences of the training data of different models. This information can guide the improvement of model training strategies, such as adjusting the data collection and preprocessing methods, to optimize the performance of LLMs in real-world applications.
\end{enumerate}

\section{Related Work}

\subsection{Chain-of-Thought Reasoning Method}
The Reasoning language model focuses on understanding and performing reasoning tasks. By introducing techniques such as chain-of-thought (CoT) and knowledge distillation, it conducts step-by-step analysis and logical reasoning for complex problems, demonstrating strong efficiency and practicality. 

At first, Wei et al.~\cite{NEURIPS2022_9d560961} investigated how generating a chain of thought, a sequence of intermediate reasoning steps, greatly enhances the ability of large language models to tackle complex reasoning tasks. It introduced a simple method called CoT prompting, where a few reasoning exemplars are provided in the prompt, enabling sufficiently large models to naturally exhibit reasoning capabilities. 
Feng et al.~\cite{NEURIPS2023_dfc310e8} explored the theoretical underpinnings of why CoT prompting significantly boosts the performance of LLMs in complex mathematical and reasoning tasks. Using circuit complexity theory, it proved that bounded-depth Transformers cannot directly solve basic arithmetic or equation problems without an impractical increase in model size, while constant-size autoregressive Transformers can effectively address these tasks by generating CoT derivations in a standard math language format.

Shao et al.~\cite{pmlr-v202-shao23a} introduced Synthetic Prompting, a technique that uses a few handcrafted examples to prompt large language models to autonomously generate additional CoT demonstrations, selecting the most effective ones to enhance reasoning performance. It employed a two-step process: a backward phase that creates clear, solvable questions matching sampled reasoning chains, followed by a forward phase that generates detailed reasoning steps for those questions, improving demonstration quality. 
Zheng et al.~\cite{NEURIPS2023_10803064} proposed a novel Duty-Distinct Chain-of-Thought (DDCoT) prompting method to enhance multimodal reasoning in LLMs by addressing challenges like labor-intensive annotation and limited flexibility in multimodal contexts. It introduced a strategy that separates reasoning into distinct duties—critical thinking via negative-space prompting for LLMs and visual recognition via integration with visual models—fostering a collaborative reasoning process.

\subsection{Language Pattern Analysis}
Analyzing language patterns allows one to identify common syntactic structures, semantic relationships, and contextual dependencies in text. This information can guide model owners in optimizing the training process of large language models.

For example, Wu et al.~\cite{wu2024comparative} investigated the reasoning patterns of OpenAI’s o1 model by comparing it with several Test-time Compute methods, including Best-of-N (BoN), Step-wise BoN, Agent Workflow, and Self-Refine, across diverse reasoning tasks in mathematics, coding, and commonsense reasoning. The authors summarized six reasoning patterns of o1 (Systematic Analysis, Method Reuse, Divide and Conquer, Self-Refinement, Context Identification, and Emphasizing Constraints) and demonstrated that these patterns are key to its enhanced reasoning capabilities. 
Hanafi et al.~\cite{hanafi-etal-2022-comparative} compared Human-in-the-Loop (HITL) systems and LLMs for pattern extraction through experiments: HITL employs IBM Watson Discovery’s pattern induction tool, iteratively generating extraction rules based on user-provided examples and feedback; GPT-3 uses various prompting strategies and post-processing techniques to extract patterns from text; the study evaluates their precision and recall across seven use cases.

Moreover, White et al. ~\cite{10.5555/3721041.3721046} proposed a method to improve interactions with LLMs like ChatGPT by introducing a catalog of prompt engineering techniques structured as reusable prompt patterns, akin to software patterns. It established a framework for documenting these patterns to address common challenges in software development tasks, iteratively applying and refining them to optimize LLM outputs and interactions.

Muñoz-Ortiz et al.~\cite{Mu_oz_Ortiz_2024} compared news texts generated by six different LLMs with human-written texts across morphological, syntactic, psychometric, and sociolinguistic dimensions. It found that human texts exhibit greater sentence length variation, richer vocabulary, and stronger emotional expressions, while LLM texts are more objective, using more numbers and pronouns. 
Sandler et al.~\cite{sandler2024linguisticcomparisonhumanchatgptgenerated} analyzed conversations generated by ChatGPT versus human dialogues using LIWC (Linguistic Inquiry and Word Count) across 118 language categories. It showed that human dialogues are more varied and authentic, while ChatGPT excels in social processes, cognitive style, and positive emotional tone, though it shows no significant difference in overall positive or negative emotional expression.

Our study primarily focuses on the differences in the reasoning processes and outputs of various reasoning language models. By comparing these distinctions, we can better understand how each model approaches problem-solving and identify strengths and limitations in their reasoning capabilities. This analysis aims to inform the development of more effective and reliable language models for complex tasks.

\section{Characteristics Analysis of Reasoning and Output}

\subsection{Experiment Settings}

\subsubsection{Reasoning Language Models and Evaluation Dataset}

\paragraph{Reasoning Language Models.}

This study compares four representative cutting-edge reasoning language models based on their core design philosophies and technical features. 

The following introduces four reasoning language models focused on long Chain-of-Thought reasoning: GPT-o1, from OpenAI, excels in mathematics and coding. DeepSeek-R1, an open-source model, enhances reasoning through self-reflection. Grok-3, by xAI, performs strongly in scientific and mathematical tasks. Kimi-k1.5, from Moonshot AI, optimizes reasoning across text, image, and code tasks.

\paragraph{Evaluation Dataset.}

This study compares four representative cutting-edge reasoning language models based on their core design philosophies and technical features. We have constructed a multi-domain evaluation dataset with the following structure, designed to rigorously assess the reasoning capabilities of advanced language models across diverse fields. This dataset is sourced from a combination of established benchmarks, including widely recognized General Reasoning open-source datasets. It encompasses eight domains: Humanities, Puzzles, Adversarial, Programming, Finance, Mathematics, Medicine, and Physics. 
To maintain consistency and depth, we hand-selected 10 representative data points for each domain, tailored to their specific selection criteria as outlined in Table \ref{tab:evaluation_dataset}, resulting in 80 carefully curated evaluation items.

\vspace{-20pt}
\begin{table}[htbp]
\centering
\caption{Structure of the Multi-Domain Evaluation Dataset}
\begin{tabular}{l|l}  
\toprule
\multicolumn{1}{c|}{\textbf{Domain}} & \multicolumn{1}{c}{\textbf{Selection Criteria}} \\
\midrule
Humanities  & Hand-selected ethical dilemmas and historical analysis questions \\
Puzzles     & Classic lateral thinking puzzles and logical paradoxes \\
Adversarial & Covers safety testing and adversarial prompt scenarios \\
Programming & Typical problems in algorithm design and code debugging \\
Finance     & Cases focused on risk assessment and investment strategy optimization \\
Mathematics & Standard multi-step proofs and equation solving questions \\
Medicine    & Simplified designs based on real diagnostic cases \\
Physics     & Fundamental questions in theoretical derivations and experimental design \\
\bottomrule
\end{tabular}
\label{tab:evaluation_dataset}
\end{table}
\vspace{-20pt}

\subsubsection{Evaluation Metrics}

This study systematically evaluates the models' reasoning processes and output quality using the following metrics.

\paragraph{Total Reflection Count ($TRC$).}  
The $TRC$ measures the cumulative number of reflection occurrences in the reasoning texts of a specific category and model. A higher $TRC$ indicates that the model frequently reflects on the data within that domain, suggesting a more complex reasoning process and a greater need for logical problem-solving. This metric highlights the depth of the model's reasoning in addressing domain-specific challenges.

\paragraph{Reflection Data Count ($RDC$).}  
The $RDC$ represents the number of reasoning texts that contain at least one reflection keyword in a specific category and model. A higher $RDC$ implies the model performs broader reflections across a larger proportion of texts in that domain. This can be interpreted as evidence that the trainer applied extensive \textit{Long Chain-of-Thought} processing during the model's training phase for this type of data. This metric underscores the breadth of the model's reasoning in the domain.

\paragraph{Consistency Score ($CS$).} 
When evaluating reasoning language models, we focus on their consistency with high-quality reference models like GPT-o1. The Consistency Score measures this. It assesses the alignment of the model's output with a given outline from multiple dimensions. The model must understand the outline's structure, cover all key points, follow the logical order, avoid irrelevant content, and adhere to logical rules. As shown in Appendix~\ref{appendix}, the score ranges from 1 to 5, with a higher score indicating better consistency. This score is automatically evaluated by Doubao-1.5-Pro using our predefined scoring rules.

\subsection{Reasoning Pattern Analysis}

\subsubsection{Self-Reflection Pattern}

From the comparative analysis of $TRC$ and $RDC$ across different models and reasoning domains, we can observe distinct patterns in the depth and breadth of reflection. The coding and math domains generally exhibit the highest reflection depth ($TRC$), suggesting that models tend to engage in more extensive step-by-step reasoning when solving problems in these areas. In Table~\ref{tab:TRC}, kimi-k1.5 demonstrates an exceptionally high $TRC$ in coding (69), significantly exceeding the other models. Similarly, Grok-3 and DeepSeek-R1 also exhibit relatively deep reasoning in coding and math, indicating a consistent trend where numerical and structured reasoning tasks necessitate deeper logical progression.

\begin{table}[ht]
\centering
\caption{$TRC$ in Reasoning Process Across Models and Domains when using GPT-o1 as Baseline Model}
\begin{tabular}{l|cccccccc}
\toprule
\textbf{Model} & {Humanities} & {Riddles} & {Advbench} & {Coding} & {Finance} & {Math} & {Medical} & {Physics} \\
\midrule
DeepSeek-R1 & 13 & 20 & 14 & 34 & 37 & 32 & 10 & 23 \\
Grok-3      & 17 & 35 & 6  & 43 & 36 & 40 & 12 & 30 \\
kimi-k1.5    & 38 & 21 & 17 & 69 & 21 & 17 & 3  & 21 \\
\bottomrule
\end{tabular}
\label{tab:TRC}
\end{table}

\begin{table}[ht]
\centering
\caption{$RDC$ in Reasoning Process Across Models and Domains when using GPT-o1 as Baseline Model}
\begin{tabular}{l|cccccccc}
\toprule
\textbf{Model} & {Humanities} & {Riddles} & {Advbench} & {Coding} & {Finance} & {Math} & {Medical} & {Physics} \\
\midrule
DeepSeek-R1 & 8  & 10 & 9  & 9 & 8  & 9  & 6 & 6 \\
Grok-3      & 6  & 7  & 4  & 7 & 8  & 10 & 4 & 9 \\
kimi-k1.5    & 7  & 6  & 10 & 9 & 9  & 8  & 3 & 7 \\
\bottomrule
\end{tabular}
\label{tab:RDC}
\end{table}

\begin{figure}[ht]
    \centering
    \includegraphics[width=0.85\textwidth]{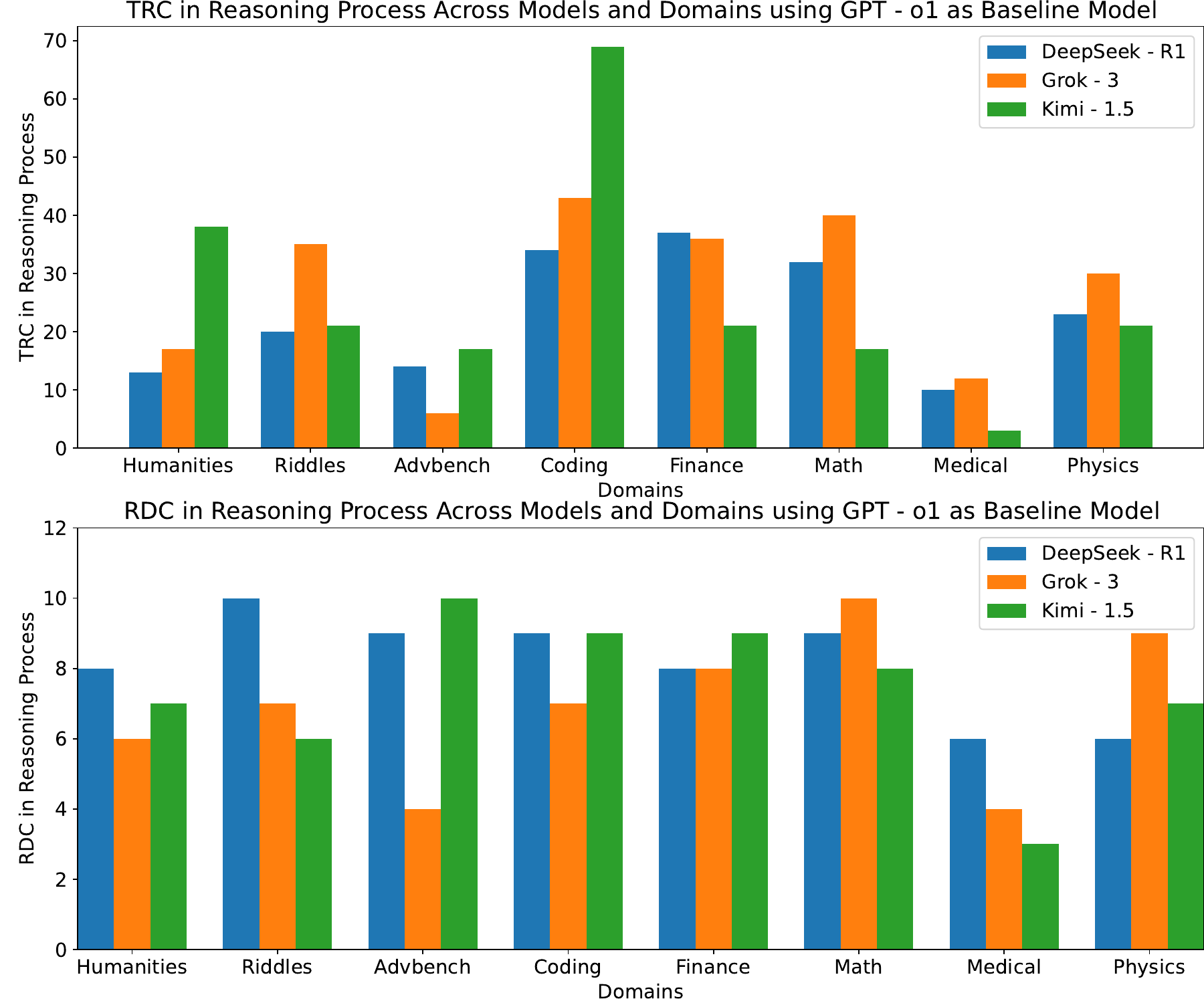}
    \caption{Comparison Analysis of $TRC$ and $RDC$ in the Reasoning Process Across Models and Domains with GPT-o1 as the Baseline Model}
    \label{fig:TRC_RDC_Com}
\end{figure}

Conversely, the breadth of reflection ($RDC$) is more evenly distributed across domains, with models generally showing higher engagement in math, coding, and finance, indicating that these domains require reasoning over a wider range of data points. However, in fields like medical and advbench, models exhibit a lower $RDC$ as shown in Table~\ref{tab:RDC}, suggesting that either their responses are more deterministic, relying on directly retrieved knowledge rather than iterative reflection, or they tend to avoid detailed responses in sensitive topics. Table~\ref{tab:TRC} and Table~\ref{tab:RDC} present the specific experimental results of ours, and Figure~\ref{fig:TRC_RDC_Com} shows the intuitive pattern evaluation results among different models.

DeepSeek-R1: This model demonstrates a balanced reasoning pattern across multiple domains. It shows substantial depth in finance with a $TRC$ of 37 and math with a $TRC$ of 32, aligning with the observation that numerical reasoning benefits from iterative logical steps. However, its $RDC$ remains moderate across all domains, with no standout figures. This indicates that while DeepSeek-R1 engages in deeper reasoning when necessary, it does not significantly expand the breadth of its reflection across different cases. Its relatively lower $TRC$ in medical, with a value of 10, and in physics, with a value of 23, suggests that it might rely more on directly retrieved knowledge in these specialized domains rather than deep deductive reasoning.

Grok-3: Unlike DeepSeek-R1, Grok-3 exhibits an interesting contrast, with an extremely high $TRC$ in riddles at $35$, indicating that it tends to explore multiple steps when solving ambiguous or wordplay-based problems. It also shows strong engagement in math with a $TRC$ of 40 and in coding with a $TRC$ of 43, further supporting the trend that computational reasoning prompts deeper reflection. However, its $RDC$ values do not strongly correlate with its $TRC$, meaning that while deeply reasons through certain domains, it does not necessarily reflect over a broad range of data points. This could imply that its knowledge retrieval mechanism is efficient enough to limit unnecessary exploration, particularly in structured problems.

kimi-k1.5: Among the three models, kimi-k1.5 exhibits the most extreme reasoning behavior, with an exceptionally high $TRC$ in coding at 69, far surpassing the other models. This suggests that it relies heavily on step-by-step logical progression in computational tasks. However, its $TRC$ for medical is the lowest at 3, likely indicating a reliance on pre-existing medical knowledge rather than engaging in extensive inference. Similarly, its relatively lower $TRC$ in finance at 21 and in physics at $21$ suggests that it does not deeply reflect on these domains, potentially due to a stronger reliance on factual recall. Interestingly, its $RDC$ in advbench is the highest at 10, suggesting that even though the model may refuse to answer sensitive questions at times, it considers a diverse range of cases when it does engage.

One notable observation is the significantly lower $TRC$ in advbench across all models. This does not necessarily indicate weaker reasoning capabilities but could result from the models refusing to engage with certain sensitive questions. Since models are trained with ethical constraints, they may generate fewer reasoning steps when encountering adversarial or policy-sensitive content. Similarly, the relatively lower $TRC$ in the medical domain suggests that models may prioritize direct factual recall over extensive reasoning, which aligns with the expectations for a field that heavily depends on authoritative knowledge rather than logical deduction.

\subsubsection{Reasoning Consistency Pattern}

\begin{figure}[ht]
    \centering
    \begin{subfigure}[b]{0.495\textwidth}
        \centering
        \includegraphics[width=\textwidth]{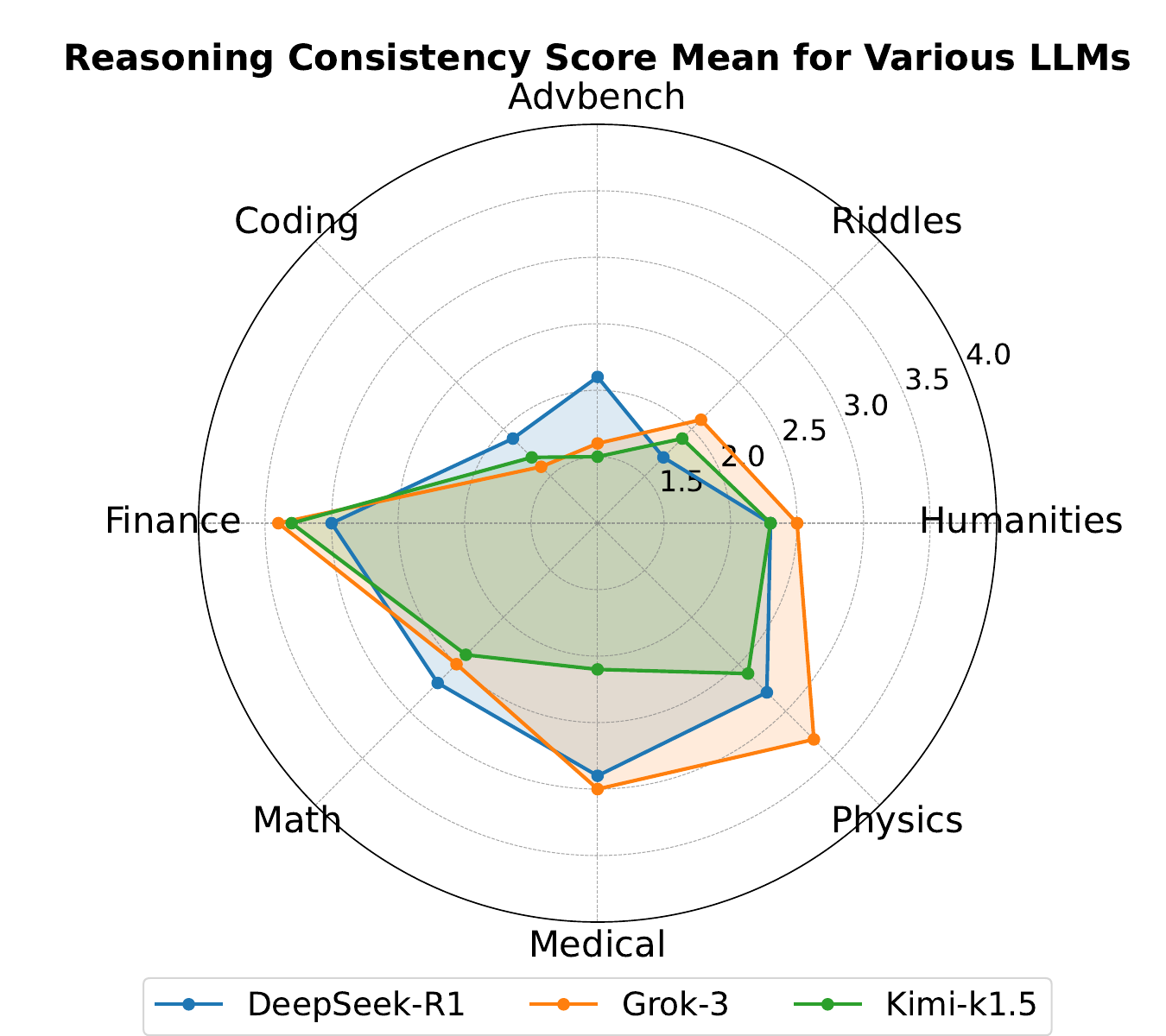}
        \caption{$CS$ Mean in Reasoning}
        \label{fig:reasoning_mean}
    \end{subfigure}
    \begin{subfigure}[b]{0.495\textwidth}
        \centering
        \includegraphics[width=\textwidth]{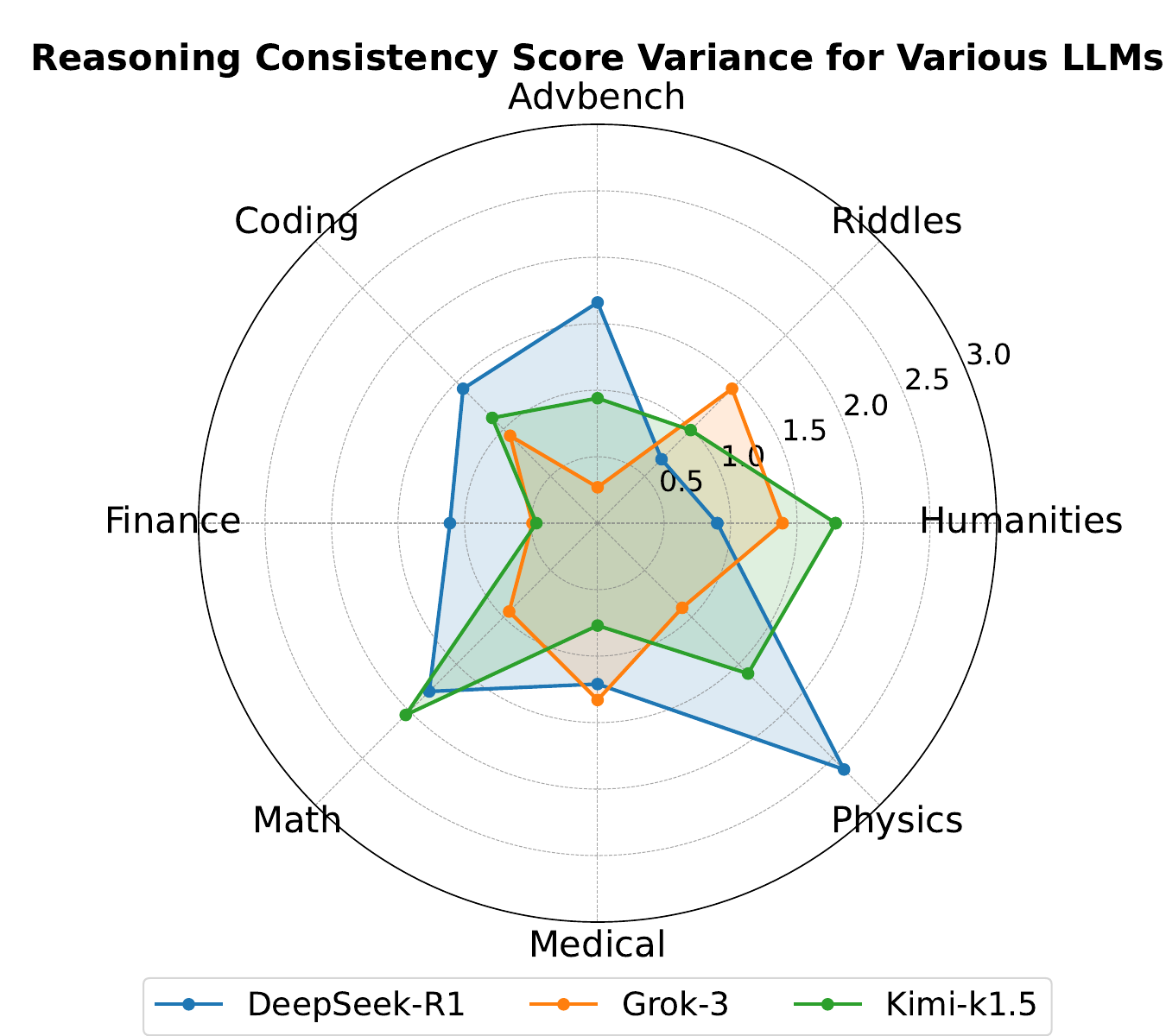}
        \caption{$CS$ Variance in Reasoning}
        \label{fig:reasoning_var}
    \end{subfigure}
    \caption{Comparison of the Mean and Variance of the Consistency Score for the Reasoning Process.}
    \label{fig:reasoning_cs}
\end{figure}

The Consistency Score evaluates how well different LLMs align with the reference model GPT-o1 across multiple domains. A higher score indicates that the model closely follows the reasoning structure of the reference model, covers all key points, and maintains logical consistency, while a lower score suggests deviations in these aspects. Figure~\ref{fig:reasoning_cs} shows the mean and variance of the $CS$ for the Reasoning Process.

As shown in Table~\ref{csr-d}, DeepSeek-R1 demonstrates varying levels of consistency across different domains. The model shows relatively stable performance in the finance and medical domains, with an average score of 3.0 and 2.9, respectively. The mode in both cases is 4.0, indicating that DeepSeek-R1 frequently aligns well with GPT-o1 in these fields. Similarly, in the math domain, the model achieves an average score of 2.7, with a median of 3.0, reflecting a moderate level of consistency. This suggests that DeepSeek-R1 performs better in structured numerical reasoning tasks.  

However, the model exhibits weaker consistency in coding, riddles, and adversarial benchmark tasks. In the coding domain, the average score is 1.9, while in riddles, it is 1.7, and in adversarial benchmark tasks, it is 2.1. These lower scores indicate that DeepSeek-R1 struggles to maintain alignment with GPT-o1 in tasks requiring complex reasoning or handling sensitive content. The variance in the physics domain is 2.62, and in math, it is 1.79, showing significant fluctuations in performance. This variability may be attributed to differences in problem complexity or the model adopting different reasoning approaches depending on the question.  

The mode distribution further highlights DeepSeek-R1's reasoning characteristics. The most frequently occurring score in finance, math, and medical domains is 4.0, suggesting that the model often follows GPT-o1’s reasoning logic in these areas. Conversely, the most frequent score in physics, coding, and riddles is 1.0, indicating a tendency to deviate from the expected reasoning structure. This suggests that in certain domains, DeepSeek-R1 relies more on heuristic methods or direct retrieval rather than structured logical inference.  

The reasoning strategy of DeepSeek-R1 differs from that of GPT-o1, leading to cases where the model demonstrates strong performance while still receiving a low Consistency Score. This suggests that DeepSeek-R1 may adopt alternative reasoning approaches that deviate from the structured, logical inference favored by GPT-o1. The observed differences in Consistency Score may also indirectly reflect disparities between the training data of DeepSeek-R1 and that of OpenAI models. These discrepancies could result in variations in reasoning patterns, affecting the model's ability to align with GPT-o1’s structured outline while still achieving high task performance.

\begin{table}[htbp]
    \centering
    \caption{Consistency Score Analysis for Different LLMs Reasoning Process}
    \begin{subtable}[b]{\textwidth}
        \centering
        \caption{Consistency Score Analysis for DeepSeek-R1 Reasoning Process}
        \begin{tabularx}{\textwidth}{@{\extracolsep{\fill}}l|cccccc@{}}
            \toprule
            Category & Max & Min & Median & Mode & Mean & Variance \\
            \midrule
            Humanities~ & 4.00 & 1.00 & 2.00 & 2.00 & 2.30 & 0.90 \\
            Riddles & 3.00 & 1.00 & 1.50 & 1.00 & 1.70 & 0.68 \\
            Advbench & 4.00 & 1.00 & 1.50 & 1.00 & 2.10 & 1.66 \\
            Coding & 4.00 & 1.00 & 1.50 & 1.00 & 1.90 & 1.43 \\
            Finance & 4.00 & 1.00 & 3.00 & 4.00 & 3.00 & 1.11 \\
            Math & 4.00 & 1.00 & 3.00 & 4.00 & 2.70 & 1.79 \\
            Medical & 4.00 & 1.00 & 3.00 & 4.00 & 2.90 & 1.21 \\
            physics & 5.00 & 1.00 & 2.50 & 1.00 & 2.80 & 2.62 \\
            \bottomrule
        \end{tabularx}
        \label{csr-d}
    \end{subtable}
    \begin{subtable}[b]{\textwidth}
        \centering
        \caption{Consistency Score Analysis for Grok-3 Reasoning Process}
        \begin{tabularx}{\textwidth}{@{\extracolsep{\fill}}l|cccccc@{}}
            \toprule
            Category & Max & Min & Median & Mode & Mean & Variance \\
            \midrule
            Humanities~ & 4.00 & 1.00 & 2.00 & 2.00 & 2.50 & 1.39 \\
            Riddles & 4.00 & 1.00 & 2.00 & 1.00 & 2.10 & 1.43 \\
            Advbench & 2.00 & 1.00 & 2.00 & 2.00 & 1.60 & 0.27 \\
            Coding & 4.00 & 1.00 & 1.00 & 1.00 & 1.60 & 0.93 \\
            Finance & 4.00 & 2.00 & 3.50 & 4.00 & 3.40 & 0.49 \\
            Math & 4.00 & 1.00 & 3.00 & 3.00 & 2.50 & 0.94 \\
            Medical & 4.00 & 1.00 & 3.50 & 4.00 & 3.00 & 1.33 \\
            Physics & 5.00 & 2.00 & 3.00 & 3.00 & 3.30 & 0.90 \\
            \bottomrule
        \end{tabularx}
        \label{csr-g}
    \end{subtable}
    \begin{subtable}[b]{\textwidth}
        \centering
        \caption{Consistency Score Analysis for Kimi-k1.5 Reasoning Process}
        \begin{tabularx}{\textwidth}{@{\extracolsep{\fill}}l|cccccc@{}}
            \toprule
            Category & Max & Min & Median & Mode & Mean & Variance \\
            \midrule
            Humanities~ & 4.00 & 1.00 & 2.00 & 1.00 & 2.30 & 1.79 \\
            Riddles & 4.00 & 1.00 & 2.00 & 1.00 & 1.90 & 0.99 \\
            Advbench & 4.00 & 1.00 & 1.00 & 1.00 & 1.50 & 0.94 \\
            Coding & 4.00 & 1.00 & 1.00 & 1.00 & 1.70 & 1.12 \\
            Finance & 4.00 & 2.00 & 3.00 & 3.00 & 3.30 & 0.46 \\
            Math & 5.00 & 1.00 & 2.50 & 1.00 & 2.40 & 2.04 \\
            Medical & 4.00 & 1.00 & 2.00 & 2.00 & 2.10 & 0.77 \\
            Physics & 4.00 & 1.00 & 3.00 & 1.00 & 2.60 & 1.60 \\
            \bottomrule
        \end{tabularx}
        \label{csr-k}
    \end{subtable}
\end{table}

Grok-3 demonstrates high consistency in structured domains such as finance, medical reasoning, and physics, with average Consistency Scores of 3.4, 3.0, and 3.3, respectively, and relatively low variance. This indicates that Grok-3's reasoning structure closely aligns with GPT-o1 in these tasks, effectively adhering to predefined logical frameworks and covering key information. This performance may be attributed to the well-defined nature of knowledge in these fields, allowing the model to generate expected outputs more reliably.

However, as shown in Table~\ref{csr-g}, in coding and adversarial reasoning (Advbench) tasks, Grok-3 exhibits significantly lower Consistency Scores, averaging 1.6, with low variance, suggesting poor consistency and unstable outputs. In particular, the most frequent score for coding tasks is 1, indicating that the model frequently fails to align well with reference reasoning. This may stem from Grok-3’s reasoning strategy, which, instead of step-by-step deduction or logical inference, may rely more heavily on pattern matching or direct retrieval of known information. As a result, its performance is weaker in tasks requiring structured reasoning.  

Additionally, in mathematics and humanities, Grok-3's Consistency Score hovers around 2.5, indicating a moderate level of alignment with GPT-o1’s reasoning approach. Notably, although the median and mode scores are relatively high in mathematics, the large variance suggests that Grok-3’s reasoning stability fluctuates significantly, potentially due to variations in problem types or input structures.  

Kimi-k1.5 demonstrates strong consistency in structured domains such as finance, where the average Consistency Score reaches 3.3 with low variance. This suggests that the model's reasoning structure aligns well with GPT-o1, effectively following predefined logical frameworks and ensuring comprehensive content coverage. Similar trends are observed in physics and mathematics, with mean scores of 2.6 and 2.4, respectively, though higher variance in mathematics indicates fluctuations in reasoning stability depending on the specific problem type.  

In contrast, Kimi-k1.5 struggles with adversarial reasoning (Advbench) and coding tasks, where the average Consistency Scores are 1.5 and 1.7, respectively. The low median and mode values and relatively low variance suggest that the model frequently exhibits inconsistent reasoning structures. This may be due to its reliance on heuristic shortcuts or pattern matching rather than step-by-step logical inference, leading to discrepancies in reasoning depth and alignment with GPT-o1.

For humanities and riddles, Kimi-k1.5 achieves moderate Consistency Scores, averaging 2.3 and 1.9, respectively. However, the high variance of 1.79 in humanities suggests significant instability in reasoning, likely influenced by the model’s sensitivity to diverse linguistic styles and abstract concepts. In medical reasoning, the mean score of 2.1 with relatively low variance implies a stable yet somewhat limited alignment with GPT-o1, indicating that while the model captures general medical knowledge, its logical structuring may lack the depth and consistency seen in more structured domains.

\subsubsection{Human Observation Pattern}

In the Humanities domain, DeepSeek-R1 frequently uses “let me think” and “let me check,” each appearing 4 times, reflecting a rigorous yet repetitive reflection that may hinder creativity. Grok-3 predominantly uses “let me think,” occurring 7 times, but also shows contradictory cues with “doesn’t make sense” appearing 2 times, indicating occasional logical leaps, while Kimi-k1.5 relies almost exclusively on “let me check,” which appears 22 times, revealing a rigid, inflexible pattern.

In the Coding domain, DeepSeek-R1’s frequent use of “Let’s think,” appearing 15 times, demonstrates clear, step-by-step reasoning and high consistency. Grok-3 balances its reflections with “let me think” occurring 16 times alongside “make sure” occurring 15 times, leading to moderate consistency, whereas Kimi-k1.5, with “Let’s think” used 28 times and “think again” used 9 times, exhibits strong logical coverage. In Mathematics, DeepSeek-R1 employs a dual-insurance mechanism by using “let me check” 12 times and “make sure” 7 times, ensuring robust logical stability. Grok-3 distributes its reflective phrases evenly, though the 6 occurrences of “makes sense” suggest a subjective judgment that may introduce ambiguity, while Kimi-k1.5 uses only a total of 17 reflective cues, resulting in insufficient logical coverage and low consistency.

\subsection{Output Pattern Analysis}

\subsubsection{Output Consistency Pattern}

\begin{figure}[ht]
    \centering
    \begin{subfigure}[b]{0.495\textwidth}
        \centering
        \includegraphics[width=\textwidth]{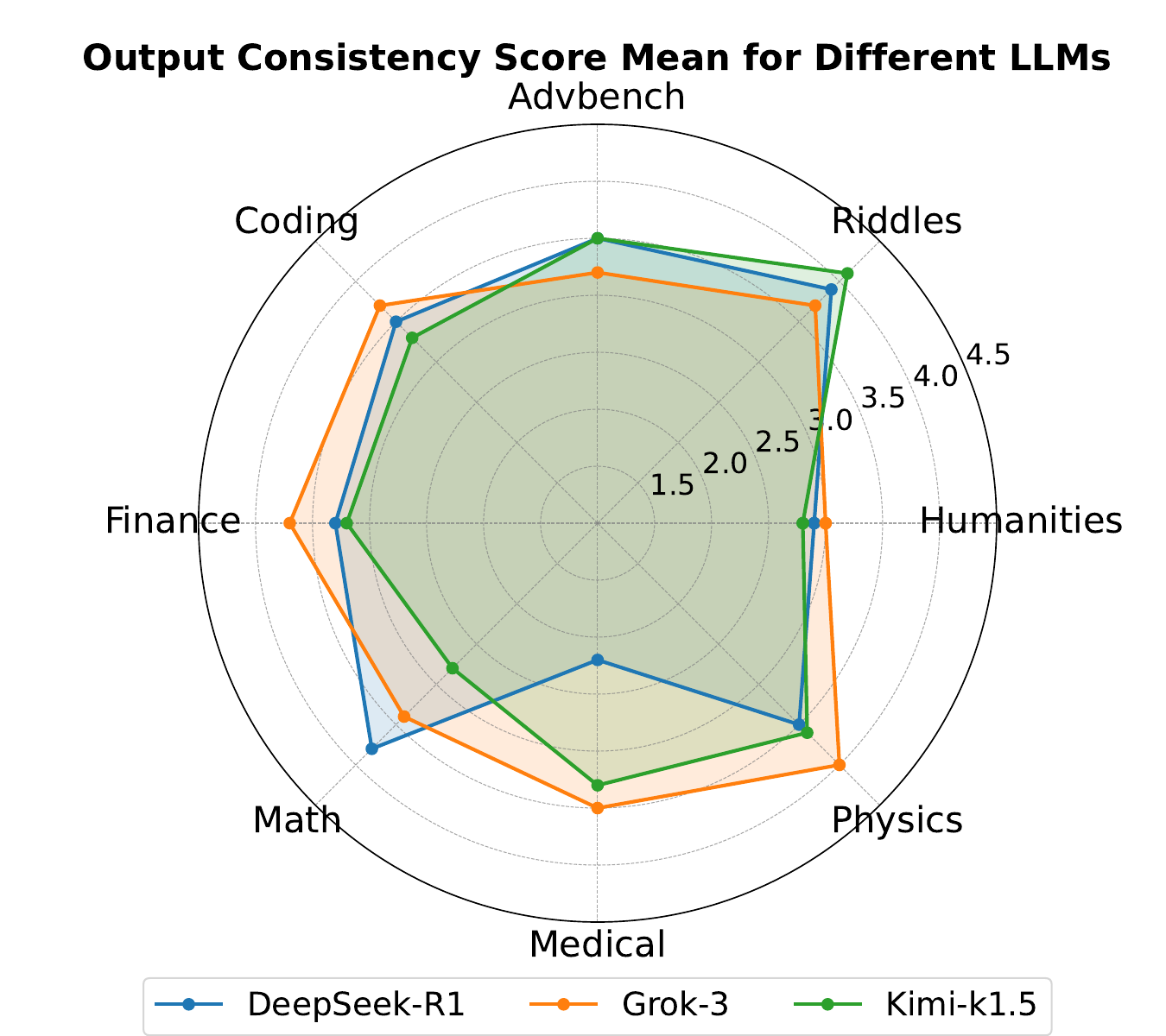}
        \caption{$CS$ Mean in Outputs}
        \label{fig:Output_mean}
    \end{subfigure}
    \begin{subfigure}[b]{0.495\textwidth}
        \centering
        \includegraphics[width=\textwidth]{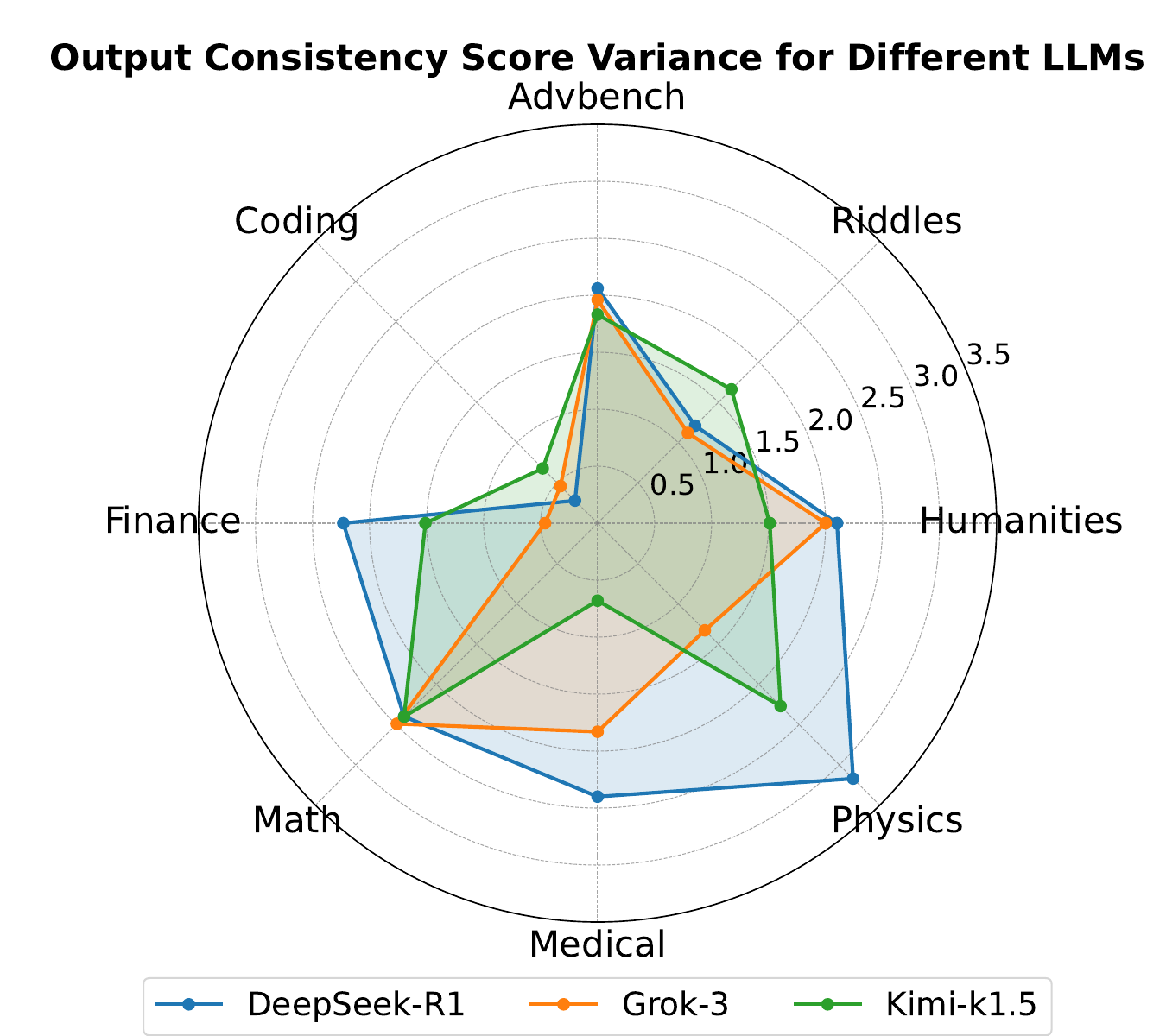}
        \caption{$CS$ Variance in Outputs}
        \label{fig:Output_var}
    \end{subfigure}
    \caption{Comparison of the mean and variance of the Consistency Score for the Outputs.}
    \label{fig:output_CS}
\end{figure}

Table \ref{cso-d} shows the output consistency between DeepSeek-R1 and GPT-o1 across different categories. In Figure~\ref{fig:Output_mean}, most categories have a mean above 3.0, with the median and mode frequently at 4.00, indicating a relatively high level of consistency. However, in Figure~\ref{fig:Output_var}, variance values reveal significant differences in certain categories.

The Coding category's variance is only 0.28, with relatively close maximum and minimum values. This suggests a high level of output consistency between DeepSeek-R1 and GPT-o1. Such consistency may indicate a substantial overlap in training data related to programming tasks or that structured problems are less sensitive to variations in training approaches.

On the other hand, the Math, Medical, and Physics categories exhibit relatively high variance values of 2.40, 2.40, and 3.17, suggesting greater output differences. In these fields, knowledge is complex and frequently updated, and different models may have been trained on distinct datasets, such as medical literature, encyclopedic knowledge, or specific math problems, leading to inconsistencies in generated responses. Similarly, physics questions often involve formula derivation and numerical calculations, where variations in the coverage of physics formulae and problem-solving approaches in the training data may result in differences in the output.

Additionally, the Finance category has a variance of 2.23, indicating a certain degree of inconsistency. This could be attributed to differences in financial text sources used for training, such as news and academic papers, which may have influenced the models’ response strategies.

\begin{table}[htbp]
    \centering
    \caption{Consistency Score Analysis for Different LLMs Outputs}
    \begin{subtable}[b]{\textwidth}
        \centering
        \caption{Consistency Score Analysis for DeepSeek-R1 Outputs}
        \begin{tabularx}{\textwidth}{@{\extracolsep{\fill}}l|cccccc@{}}
            \toprule
            Category & Max & Min & Median & Mode & Mean & Variance \\
            \midrule
            Humanities & 4.00 & 1.00 & 4.00 & 4.00 & 2.90 & 2.10 \\
            Riddles & 5.00 & 1.00 & 4.00 & 4.00 & 3.90 & 1.21 \\
            Advbench & 5.00 & 1.00 & 4.00 & 4.00 & 3.50 & 2.06 \\
            Coding & 4.00 & 3.00 & 3.50 & 3.00 & 3.50 & 0.28 \\
            Finance & 5.00 & 1.00 & 4.00 & 4.00 & 3.30 & 2.23 \\
            Math & 5.00 & 1.00 & 4.00 & 4.00 & 3.80 & 2.40 \\
            Medical & 4.00 & 1.00 & 1.00 & 1.00 & 2.20 & 2.40 \\
            Physics & 5.00 & 1.00 & 4.00 & 5.00 & 3.50 & 3.17 \\
            \bottomrule
        \end{tabularx}
        \label{cso-d}
    \end{subtable}
    \begin{subtable}[b]{\textwidth}
        \centering
        \caption{Consistency Score Analysis for Grok-3 Outputs}
        \begin{tabularx}{\textwidth}{@{\extracolsep{\fill}}l|cccccc@{}}
            \toprule
            Category & Max & Min & Median & Mode & Mean & Variance \\
            \midrule
            Humanities & 4.00 & 1.00 & 4.00 & 4.00 & 3.00 & 2.00 \\
            Riddles & 5.00 & 1.00 & 4.00 & 4.00 & 3.70 & 1.12 \\
            Advbench & 5.00 & 1.00 & 4.00 & 4.00 & 3.20 & 1.96 \\
            Coding & 4.00 & 2.00 & 4.00 & 4.00 & 3.70 & 0.46 \\
            Finance & 4.00 & 2.00 & 4.00 & 4.00 & 3.70 & 0.46 \\
            Math & 5.00 & 1.00 & 4.00 & 4.00 & 3.40 & 2.49 \\
            Medical & 5.00 & 1.00 & 4.00 & 4.00 & 3.50 & 1.83 \\
            Physics & 5.00 & 1.00 & 4.00 & 4.00 & 4.00 & 1.33 \\
            \bottomrule
        \end{tabularx}
        \label{cso-g}
    \end{subtable}
    \begin{subtable}[b]{\textwidth}
        \centering
        \caption{Consistency Score Analysis for Kimi-k1.5 Outputs}
        \begin{tabularx}{\textwidth}{@{\extracolsep{\fill}}l|cccccc@{}}
            \toprule
            Category & Max & Min & Median & Mode & Mean & Variance \\
            \midrule
            Humanities & 4.00 & 1.00 & 3.00 & 4.00 & 2.80 & 1.51 \\
            Riddles & 5.00 & 1.00 & 4.50 & 5.00 & 4.10 & 1.66 \\
            Advbench & 5.00 & 1.00 & 4.00 & 4.00 & 3.50 & 1.83 \\
            Coding & 4.00 & 2.00 & 3.50 & 4.00 & 3.30 & 0.68 \\
            Finance & 4.00 & 1.00 & 4.00 & 4.00 & 3.20 & 1.51 \\
            Math & 5.00 & 1.00 & 3.00 & 1.00 & 2.80 & 2.40 \\
            Medical & 4.00 & 2.00 & 3.50 & 4.00 & 3.30 & 0.68 \\
            Physics & 5.00 & 1.00 & 4.00 & 4.00 & 3.60 & 2.27 \\
            \bottomrule
        \end{tabularx}
        \label{cso-k}
    \end{subtable}
\end{table}

Grok-3 shows solid consistency in its outputs across multiple categories. As shown in Table~\ref{cso-g}, the Physics category stands out with a mean of 4.00 and low variance (1.33), indicating that Grok-3 provides consistent outputs that align closely with expectations. The Coding and Finance categories also perform well, with mean consistency scores of 3.70 and low variance (0.46), suggesting that Grok-3's outputs are reliable in these areas.

However, Grok-3 shows more variability in other categories, such as Advbench and Medical. While the mean scores are still relatively high (3.20 and 3.50, respectively), the variance is higher, indicating that the output can vary considerably in these domains. The highest variability is observed in the Math category, where the mean score is 3.40, and the variance is 2.49, suggesting that the model struggles with generating consistent outputs for complex or abstract mathematical problems.

As shown in Table~\ref{cso-k}, kimi-k1.5 demonstrates strong consistency in its output for Riddles, with a mean score of 4.10 and a mode of 5.00, indicating that the model's outputs in this category are highly aligned with expected results. The Physics and Advbench categories also show solid performance, with mean consistency scores of 3.60 and 3.50, respectively, though the variance in Physics (2.27) suggests some occasional inconsistencies in the model's output.

In categories like Finance and Medical, kimi-k1.5 shows moderate consistency, with mean scores of 3.20 and 3.30, respectively, and relatively low variance (1.51 and 0.68). However, the model's output consistency is less stable in the Humanities and Math categories, where the mean consistency scores are 2.80 and 2.80, with variances of 1.51 and 2.40, respectively. This indicates that kimi-k1.5 may face challenges when generating outputs for more abstract or diverse questions in these areas.

\subsubsection{Human Observation Pattern}

DeepSeek-R1 delivers detailed and structured responses to clear queries, particularly excelling in Finance and Riddles with high scores; however, when faced with complex or factually precise questions—especially in Humanities and Riddles—it can exhibit misunderstandings or logical inconsistencies. Grok-3 produces relevant and logically coherent responses, performing strongly in Riddles and Finance, but may sometimes misinterpret key points or include excessive details in Humanities and some Finance queries, leading to errors in complex or abstract problems. Kimi-k1.5 generally covers key points across various categories with high alignment in Riddles and Finance, yet it occasionally deviates in Humanities and Finance by misinterpreting core questions or introducing unexpected information, particularly when dealing with complex issues. The performance of these models depends on the clarity and familiarity of the query, resulting in fluctuations in both accuracy and depth.

\section{Conclusion}
This study systematically analyzed different LLMs' reasoning processes and outputs, revealing key patterns in self-reflection, thought consistency, and human-observed reasoning. Our findings highlight significant differences in how models adjust their reasoning strategies and align with benchmark behaviors like GPT-o1. Additionally, we observed that variations in internal reasoning directly impact output accuracy, structure, and consistency. By linking internal thought processes to final outputs, our research provides a new perspective on evaluating LLMs. These insights enhance our understanding of model reasoning and offer practical guidance for improving training strategies. Future work can extend this framework to more models and tasks, further advancing the development of robust and interpretable LLMs.

\appendix
\section{Appendix}
\label{appendix}

\begin{tcolorbox}[
  breakable,
  colback=gray!10,    
  colframe=black,     
  coltitle=black,     
  colbacktitle=gray!50,  
  title=Judge System Prompt,     
  boxrule=0.8pt,      
  fonttitle=\small\bfseries 
]
\small
Below are the evaluation criteria for assessing the alignment of a model's thinking process with a given outline. These criteria ensure that the model's output adheres to the logic and structure defined in the outline, providing a meaningful and consistent response.\\
Evaluation Criteria - Alignment of Model's Thinking Process with Outline:\\
Understanding of Outline Structure: The model should accurately comprehend the structure and hierarchy of the outline, including the main points and sub-points.\\
1. Coverage of Outline Points: The model's response should comprehensively address all the key points outlined in the provided structure.\\
2. Logical Flow and Coherence: The model's response should present the information in a coherent manner, following the logical sequence defined by the outline.\\
3. Avoidance of Irrelevant Content: The model should avoid including information that deviates from or is irrelevant to the outline's focus.\\
4. Adherence to Logical Guidelines: The model should follow logical guidelines such as avoiding contradictions, ensuring consistency, and maintaining clarity in the response.\\
Now, I will provide you with a user outline and the model's response to that outline. Please review the model's response in light of the evaluation criteria:
\end{tcolorbox}

\begin{tcolorbox}[
  breakable,
  colback=gray!10,    
  colframe=black,     
  coltitle=black,     
  colbacktitle=gray!50,  
  title=Judge Rules Prompt,     
  boxrule=0.8pt,      
  fonttitle=\small\bfseries 
]
\small
\#\#\# GPT Outline: 
The following reasoning is extracted from GPT-o1 and serves as the benchmark for evaluating other models:\\

\{The reasoning process of GPT-o1 here.\}\\

\#\#\# Model Thinking: \{The reasoning process of evaluated model here.\} \\

Use the scoring rules below to score the model’s response to the GPT outline on a scale of 1 to 5:\\
Scoring Rules:\\

Score 5: Ideal matching (5)\\
The model's response perfectly aligns with the outline, covering all points with a full understanding of the structure and hierarchy.\\
The logical flow and coherence are exactly as expected, with no irrelevant content and complete adherence to logical guidelines.\\

Score 4: Mostly matching (4)\\
The model's response largely aligns with the outline, covering most key points and showing a substantial understanding of the structure.\\
The logical flow and coherence are mostly as expected, with minimal irrelevant content and largely adherence to logical guidelines.\\

Score 3: Substantially matching (3)\\
The model's response aligns with some parts of the outline, covering some key points but shows partial understanding of the structure.\\
The logical flow and coherence are partially as expected, with some irrelevant content and partial adherence to logical guidelines.\\

Score 2: Barely matching (2)\\
The model's response barely aligns with the outline, covering few key points and showing limited understanding of the structure.\\
The logical flow and coherence are barely as expected, with significant irrelevant content and limited adherence to logical guidelines.\\

Score 1: Not matching (1)\\
The model's response does not align with the outline, missing key points and showing no understanding of the structure.\\
The logical flow and coherence are not as expected, with substantial irrelevant content and no adherence to logical guidelines.\\
Output your evaluation in the following format:\\

\# thereason: your analysis here\\

Note: this should be step-by-step analysis following the steps:\\
(a) Give an overview of the user's outline and the model's response.\\
(b) Evaluate the degree to which the model's response aligns with the user's expressed outline.\\
(c) Examine the user's outline and the model's reply respectively to determine if they contain any logical inadequacies or irrelevant content.\\

Finally, evaluate the degree of the model's adherence to the defined logical guidelines.\\

\# thescore: your score here.
\end{tcolorbox}

\begin{credits}
\subsubsection{\ackname} This research is partially supported by NSFC-FDCT under its Joint Scientific Research Project Fund (Grant No. 0051/2022/AFJ)

\subsubsection{\discintname}
The authors have no competing interests to declare that are
relevant to the content of this article.

\end{credits}
%
%
\bibliographystyle{splncs04}
\bibliography{cite}

\end{document}